\documentclass[a4paper,11pt,twocolumn,twoside]{article}
\usepackage{multirow}
\usepackage{float}
\usepackage{graphicx}
\usepackage{sepln}
\usepackage{sepln_en}
\usepackage{fullname_esp}
\usepackage{lipsum}
\usepackage[utf8]{inputenc}
\usepackage{caption}
\DeclareCaptionLabelSeparator{colon}{: }
\usepackage[T1]{fontenc}
\usepackage{CJKutf8}

\usepackage{inputenc}
\usepackage[english,spanish,es-nosectiondot, es-noindentfirst, es-nolists]{babel}

\input epsf

\title{Gradable ChatGPT Translation Evaluation}

\author {\textbf{Hui Jiao,$^1$} \textbf{Bei Peng,$^2$} \textbf{Lu Zong,$^1$} \textbf{Xiaojun Zhang,$^{1,4}$} \textbf{Xinwei Li$^3$}\\
$^1$Xi'an Jiaotong-Liverpool University\\
$^2$University of Liverpool\\
$^3$Southeast University\\
$^4$Guangdong Provincial Key Laboratory of Novel Security Intelligence Technologies\\
Xiaojun.Zhang01@xjtlu.edu.cn\\
}

\seplntranstitle{Evaluación de traducción graduada de ChatGPT}

\seplnkey{Translation prompt, ChatGPT, T3S taxonomy, Evaluation.}

\seplnabstract{ChatGPT, as a language model based on large-scale pre-training, has exerted a profound influence on the domain of machine translation. In ChatGPT, a ``Prompt'' refers to a segment of text or instruction employed to steer the model towards generating a specific category of response. The design of the translation prompt emerges as a key aspect that can wield influence over factors such as the style, precision and accuracy of the translation to a certain extent. However, there is a lack of a common standard and methodology on how to design and select a translation prompt. Accordingly, this paper proposes a generic taxonomy, which defines gradable translation prompts in terms of expression type, translation style, Part-of-Speech information and explicit statement, thus facilitating the construction of prompts endowed with distinct attributes tailored for various translation tasks. Specific experiments and cases are selected to validate and illustrate the effectiveness of the method.}

\seplnclave{Prompt de traducción, ChatGPT, Taxonomía T3S, Evaluación.}

\seplnresumen{ChatGPT, un modelo de lenguaje basado en un pre-entrenamiento a gran escala que ha tenido un profundo impacto en la traducción automática. En este contexto, un "prompt" se refiere a un segmento de texto o instrucción utilizada para dirigir el modelo hacia la generación de una respuesta específica. El diseño del prompt de traducción es crucial y puede influir en aspectos como el estilo, la precisión y la exactitud de la traducción. Sin embargo, actualmente carecemos de un estándar y metodología común para diseñar y seleccionar prompts de traducción. Por lo tanto, este artículo propone una taxonomía genérica que define prompts de traducción evaluables en términos de expresión, estilo, Part-Of-Speech (POS) y declaración explícita, facilitando la construcción de prompts con distintos atributos adaptados a diversas tareas de traducción. Se han seleccionado experimentos y casos específicos para validar e ilustrar la eficacia del método.}

\firstpageno{1}

\begin{document}

% \captionsetup[figure]{labelformat={default},labelsep=period,name={Figure.}}
\captionsetup[figure]{font={footnotesize},labelfont={default},labelformat={default},labelsep=period,name={Figure}}
\captionsetup[table]{font={footnotesize},labelfont={default},labelformat={default},labelsep=period,name={Table}}

% la siguiente instrucción sólo se debe usar si el abstract sobrescribe el texto
% la longitud variará según se necesite

\setlength\titlebox{20cm} % se aumenta el tamaño del espacio reservado para datos de título

\label{firstpage} \maketitle
\let\thefootnote\relax\footnotetext{Corresponding author: Xiaojun Zhang}
%\begin{abstract}
%Resumen del artículo con una sangría a izquierda y derecha de 0.32
%cm, justificado por ambos lados, con tamaño de fuente 11.
%
%\end{abstract}

\section{Introduction }

Machine translation (MT), one of the oldest branches of research in the field of natural language processing, involves techniques for transforming one natural language into another. As a key research area within the field of Artificial Intelligence, MT has been widely used in a wide range of fields and has attracted extensive attention from both academia and industry \cite{yang2020survey}. In recent years, there has been a growing trend towards the use of large-scale pre-trained language models for natural language processing (NLP) \cite{yang2020survey,brown2020language,amplayo2022attribute}. Large Language Models (LLMs), such as GPT-3 \cite{brown2020language}, PaLM \cite{chowdhery2022palm}, and LLaMA \cite{touvron2023llama}, which are usually trained on massive textual data, are able to capture abundant representations of the input text, and have greatly contributed to the development of MT technology \cite{lewis2019bart}.

NLP tasks have recently been significantly influenced by the emergence of ChatGPT, a powerful pre-trained LLM developed by OpenAI. The model has been trained to perform a large number of human-like tasks (e.g., question answering, code debugging, generating evaluations, etc.) with human feedback. However, the overall performance of ChatGPT on translation tasks using simple prompts and basic settings is not as good as the commercial translation products such as Google Translate and Microsoft Translate \cite{he2022tencent,zhou2022large,jiao2023chatgpt,hendy2023good}but is promising to surpass them with complex and explicit prompts. This is due to the fact that for a given complex translation task, significant differences in the performance of LLMs arise when different types/styles/levels of detail of prompts are introduced \cite{santu2023teler}.

The effective use of LLMs requires elaborate prompt engineering, which refers to the process of designing and refining the prompts or instructions provided to a large language model \cite{zhou2022large,liu2023pre}. For translation tasks of different styles, the relevant prompts that the user can try while performing the task are different. In addition, the number of details included in the prompts also largely affects the performance of LLMs when dealing with target complex translation tasks \cite{santu2023teler}. By providing clear and well-structured prompts, users can help guide LLMs in the right direction and reduce potential biases or errors. To achieve this, we propose a unified gradable prompting taxonomy for ChatGPT translation called T3S, which employs standardized criteria to categorize various types of translation prompts, thereby further enhancing ChatGPT's translation capabilities.

On the one hand, ChatGPT is utilized to conduct very broad and various types of translation tasks from general text translation to highly specialised domains (e.g., medical, legal, technical, etc.), and categorising prompts will help ChatGPT distinguish between different prompts and implement different translation strategies to meet translation needs in different domains and contexts. On the other hand, the proposed T3S taxonomy can help researchers explore in depth the working principles of LLMs and their performance differences in different domains, thus promoting further research and innovation in LLMs. Specifically, the widespread adoption of this taxonomy can potentially promote a more accurate performance assessment of ChatGPT in different translation tasks, thereby identifying specific problems that may exist in the model, providing targeted feedback for improvement and guiding the direction of model optimisation, and achieving continuous progress of the model.

\section{Related Work}

A prompt is a set of instructions provided to an LLM that enhances the functionality of the LLM by customising it \cite{liu2023pre}. In the field of large language modelling, complex tasks refer to those involving multiple steps or subtasks that require a higher level of semantic understanding, planning, reasoning, and natural language generation capabilities, which makes prompt engineering particularly critical and challenging \cite{tan-etal-2022-msp}.

In recent years, many researchers have proposed different approaches to engineering prompts. For example, one of the best-known (and easiest to implement) prompt engineering techniques is to add ``Think step by step'' to the end of the prompt. Adding this phrase improves the accuracy of the GPT-3 (text-DaVinci-002 model) across multiple tasks \cite{wei2022chain}. Moreover, \namecite{brown2020language} presented a standard question-answer pair prompting technique which produces a few-shot effect. By providing suitable output instances, LLMs are more likely to produce the desired output \cite{zhao2021calibrate}. Similar to ``Think step by step'', the Chain of Thought (CoT) prompting method guides LLMs to break down a complex task into multiple intermediate steps \cite{wang-etal-2023-towards}. \namecite{fu2022complexity}showed that separating each step with a new line in exemplar reasoning is much more effective than separating each step with a full stop. Researchers have also explored other prompt design techniques such as Reasoning and Acting (ReAct) \cite{yao2022react}, which overcomes the illusions and error propagation problems prevalent in CoT reasoning by interacting with a simple Wikipedia API. Other techniques such as Zero-shot-CoT \cite{kojima2022large} and Self-Ask \cite{press2022measuring}, improve LLMs' reasoning and action in solving questions and answering tasks. Meanwhile, \namecite{kim2023language} suggested that recursively criticising and improving its output (RCI) is superior to CoT prompts in terms of its effectiveness in reasoning ability in a range of natural language reasoning tasks.

More specifically, conversational LLMs, such as ChatGPT, have generated considerable research interest in a range of domains, with tasks ranging from answering questions for medical licensing exams to generating code snippets \cite{gilson2023does}. Correct prompt engineering has become a key skill for users wishing to utilise the full potential of ChatGPT and obtain optimal results in a variety of applications. Accordingly, there has been an influx of prompt engineering research in many different areas. \namecite{thirunavukarasu2023large} explored how prompts can be used to enhance the efficiency and effectiveness of ChatGPT in medical clinics, education, and research, while \namecite{trautmann2022legal} proposed zero-sample legal prompts engineering (LPE) to guide and enhance LLMs in natural legal language processing (NLLP) capabilities. These studies focused on NLP tasks rather than MT. In the field of machine translation, however, previous studies \cite{liu-etal-2019-incorporating-word,guo2020incorporating} have shown that while LLMs can enhance a translation system's understanding of the source text, improving its generative capabilities is more difficult. A well-developed translation system requires strong language comprehension and generation capabilities to achieve accurate and fluent translation results. Although some studies \cite{brown2020language,chowdhery2022palm} have explored the effects of different prompts on translation results, there is still a lack of systematic research on how to improve MT using prompts. With the popularity of LLM-based prompting approaches, researchers are starting to recognise the importance of introducing prompts into neural machine translation (NMT) \cite{li-etal-2022-prompt,tan-etal-2022-msp,wei2022chain}. Nevertheless, these approaches still rely on pre-training or fine-tuning the models, rather than directly applying them to “frozen” LLMs. Therefore, it is critical to study how to make the most effective use of these prompts in order to balance language comprehension and generation capabilities and achieve better results in MT.

To summarise, prompt engineering is a crucial step for the effective utilisation of LLMs. However, existing works have mainly emphasised the use of diversified prompts to improve the ability of LLMs to perform general-purpose NLP tasks, while the specification of prompts for machine translation remains under-explored. Therefore, this paper aims to fill the research gap by providing translators with a systematic approach to selecting and designing prompts, which can improve the consistency, reliability and quality of ChatGPT translations, as well as promote the development and innovation in the field of MT.

\section{ChatGPT Translation Prompting Taxonomy Design}

Through training, ChatGPT is capable of generating appropriate responses based on the given prompts. This attribute determines its high sensitivity to the information provided by the prompts. Differences in several key factors of the prompts can have a significant impact on the accuracy and performance of large language models in translation tasks \cite{santu2023teler}. These key factors are described below, where the prompt is defined by the combination of the instruction (intended target) and the source text provided to ChatGPT for performing the translation task. The prompts present differences in the instruction section under the assumption that the source text has been provided.

Ambiguous or contextually inadequate prompts can easily confuse LLMs, leading to inaccurate or irrelevant responses from these models \cite{jiangetal2020}. Therefore, providing clear and specific instructions in the prompts can help guide ChatGPT to generate the desired translation results. In the MT domain, detailed prompts usually consist of elaborating on various aspects of the translation task specification. In the following, we will delve into two basic aspects of translation task specification.

\textbf{1) Explicit Descriptions.} In translation task prompts, a clear task description is essential for obtaining accurate and relevant translation results. Specifying the translation objectives clearly and asking ChatGPT to proofread before generating the translation results helps to guide ChatGPT to a full understanding of the translation task and maintain a consistent focus throughout the translation process. This ensures a higher level of accuracy, relevance and fluency in the translation results, thereby increasing the probability of obtaining the desired translation outcome.

\textbf{2) Contextual Information.} Contextual information plays a key role in enhancing ChatGPT's understanding of words, phrases, and sentences in the source text, thus helping to reduce ambiguity and misunderstanding. Models can optimise translation decisions based on context, so as to avoid translation errors \cite{popescu2019context}. For example, for polysemous words, context can help ChatGPT determine the correct word meaning. This is crucial for producing accurate translations, as the same word may be translated differently in different contexts. Additionally, contextual background helps to maintain consistency in the model's expression throughout the translation process. When the source text has certain specific usages or terms in the context, the model can retain them in the translation based on the context, ensuring a coherent translation.

Translation prompts consist of instructions and source text. Assuming that the source text is invariant in a given translation task, we believe that the difference between prompts lies in the explicit descriptions and contextual information they contain. To address the key factors of prompt selection and design, we will classify the prompts for ChatGPT translation tasks according to the following four aspects.

\subsection{Expression Types}

There are two main types of expressing translation prompts: the single-turn prompt and the multiple-turn prompt. Both types of expression can be effective in different contexts. Single-turn prompts involve presenting the model with a solitary input, typically in the form of a single sentence or a brief textual segment (``\textit{Please translate the following text...}''), for the model to translate into the target language. Whereas multiple-turn prompts incorporate conversational interactions, usually consisting of multiple dialogue rounds, so that the model can better understand and perform the translation task. For example, a user can ask ChatGPT to check and revise the translation after it responds to the first round of prompts (``\textit{Please translate it again / Please revise the translation}''). Such prompts can be used for more complex translation tasks, where context, clarification, follow-up questions, etc., may need to be taken into account. Compared to single-turn prompts, the use of multiple-turn prompts can significantly enhance the comprehension of ChatGPT, and reduce its tendency to generate irrelevant or inaccurate responses \cite{pan2023preliminary}. Depending on the nature and needs of the translation task, choosing the appropriate prompt expression form can help ChatGPT generate the required translation output more accurately.

\subsection{Translation Style}

Depending on the translation task in different translation domains to deal with different genres such as literary, medicine, legal and commercial texts, the desired prompt needs to be defined and selected according to the translation style. Defining translation style involves determining the content and expression to be conveyed in a translation task. It relates to the specific translation target and audience. Encompassing facets such as affective undertones, tonal modulations, and the amplitude of linguistic exposition, the election of a fitting translation style exerts a discernible impact on the ultimate rendition. Thus, the inclusion of relevant contextual information such as target audience and level of expression in the prompt can provide ChatGPT with additional information to produce a more accurate translation result.

\subsection{POS Information}

Part of Speech (POS) is a grammatical category that includes nouns, verbs, adjectives, adverbs, etc. \cite{hlaing2022improving}. Many NLP tasks benefit from the use of POS tags. In translation tasks, POS tags help ChatGPT to capture the grammatical structure of a sentence in the source language, and accurately locate the grammatical roles of each word in the sentence, so as to better translate it into the grammatical structure of the target language, eliminating word ambiguities, and further enhancing the natural fluency of the translation result. For instance, \namecite{fengetal2020} demonstrated that incorporating POS tagging information into the target side can significantly improve the translation performance of the NMT system in both Chinese-to-English and German-to-English translation pairs. Further, \namecite{hlaing2022improving} conducted an NMT study using POS tagging information on low-resource language pairs, explicitly pointing out the necessity of integrating POS tags when using NMT models that include linguistic features. 

\subsection{Few-shot Prompts}

LLMs can benefit from example-based learning, which involves providing specific input-output pair examples (a small number of examples). This can help models to better understand task requirements and generate appropriate output \cite{brown2020language}. We believe that including a few input-output examples in the prompt will improve the performance of the LLMs without any adjustments to the parameters or architecture. For example, by including specific terminology and styles in the few-shot prompts, ChatGPT is able to adapt quickly across different translation domains, generating domain-appropriate translations.

We propose a gradable prompting taxonomy for ChatGPT translation which is categorised into five different levels based on the above four key elements in prompt design including expression type, translation style, POS information, and few-shot prompts. We named it T3S standing for expression type, translation style, POS information and few-shot 5-level prompting taxonomy.

\begin{figure*}[!ht]
\begin{center}
\includegraphics[width=1\linewidth]{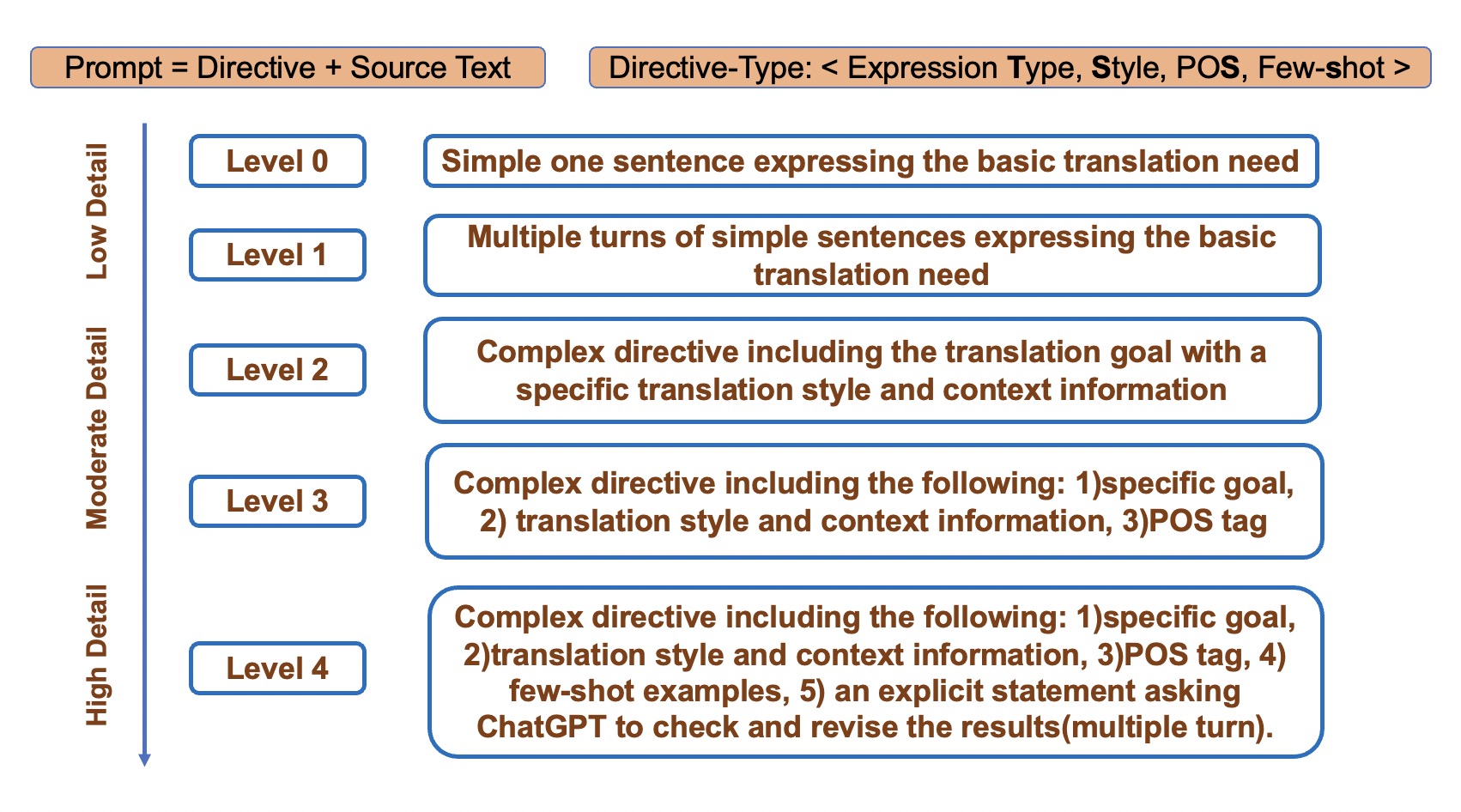} 
\caption{T3S Taxonomy.}
\label{fig.1}
\end{center}
\end{figure*}

More precisely, level ``0'' represents the lowest level of detail, where only basic translation is required with the general prompt such as ``\textit{Please translate the following text…}''; Level ``1'' distinguishes the expression type of the single-turn form with the multiple-turn form; Level ``2'' adds translation style instruction providing contextual information; Level ``3'' integrates POS information into translation instruction; and Level ``4'' represents the highest level of detail, where the multiple-turn prompts include clear instructions, explicit contextual information of translation with few-shot examples, and an explicit statement asking ChatGPT to check and revise the results.

\section{Experimental Validation}

In order to verify the rationality and validity of the T3S taxonomy, we set up an evaluation experiment. In the following, we show the details of the experimental setup, including the adopted dataset and evaluation metrics. The results and analyses of the experiment are also presented.

\subsection{Dataset}
We evaluated the translation quality of ChatGPT at different levels of prompt on the Flores-101 \cite{goyal2022flores}dataset. The dataset consists of 1012 sentences extracted from the English Wikipedia covering a wide variety of topics and domains. In real translation applications, ChatGPT needs to process texts from a variety of topics and domains. A dataset covering different topic domains can help us evaluate ChatGPT's generalisation ability and gain a more comprehensive understanding of ChatGPT's translation performance under different contexts. Moreover, these sentences have been translated into 101 languages by professional translators through a rigorously controlled process with automated and manual quality checks. Furthermore, all translations are multilingual aligned. Such a high-quality and high-coverage dataset ensures the accuracy and consistency of reference translations, and better helps us understand and evaluate the quality and performance of ChatGPT's translations. However, a graded translation quality assessment for multiple languages may add complexity and resource requirements. Therefore, for a clearer direct comparison of translation quality and to save time and cost, we only used the Chinese-English bilingual corpus for the assessment. 

\subsection{Metrics}
To assess translation quality at all levels, we employed the most commonly used BLEU score \cite{papineni2002bleu}. Additionally, we utilised CHrF \cite{popovic-2015-chrf}, TER \cite{snover-etal-2006-study} and ROUGE-1, ROUGE-2, and ROUGE-L \cite{lin-2004-rouge}, and calculated the F1 average of the scores of the ROUGE series to provide a more comprehensive assessment of translation quality at all levels. This is because one indicator may be more sensitive to certain aspects of translation quality, while another may capture different aspects of quality.

\subsection{Prompt Construction}
Prompts for each level were meticulously crafted to align with the taxonomy's gradable elements. Concretely, regarding the translation styles in Level 2, we added the labels (domain \& topic) in the dataset as translation styles to the prompt, such as wikinews(business), wikivoyage(travel), wikivoyage(sports), wikibooks(sociology/culture), etc. As for the POS information, we used the open-source natural language processing tool spaCy as our lexical annotation tool to preprocess the source text. With respect to the few-shot examples in Level 4, we randomly selected two sets of source text and target text pairs under the same domain and topic as examples to guide ChatGPT for translation.

\captionsetup[table]{labelsep=colon}
\begin{table}
\renewcommand{\arraystretch}{1.5} % 调整行高
\resizebox{\columnwidth}{!}{%
\begin{tabular}{|c|c|c|c|c|}
\hline
Translation & BLEU  & CHrF & ROUGE F1(avg) & TER   \\ \hline
Level 0     & 38.42 & 30.77 & 0.6132         & 160.19 \\ \hline
Level 1     & 38.93 & 31.07 & 0.6165         & 146.90 \\ \hline
Level 2     & 40.25 & 32.41 & 0.6256         & 117.12 \\ \hline
Level 3     & 41.25 & 33.57 & 0.6303         & 122.78 \\ \hline
Level 4     & 42.88 & 36.24 & 0.6523         & 112.78 \\ \hline
\end{tabular}%
}
\caption{ChatGPT's Translation Performance at All Levels.}
\label{Table 1}
\end{table}

\subsection{Results}
The experimental results, as shown in Table \ref{Table 1}, indicate that the translation quality of ChatGPT improves accordingly as the prompt level increases. Specifically, ChatGPT obtains a BLEU score of 38.42 for the basic translation prompt at level 0. This level of prompt is only the most basic translation requirement and does not contain any additional contextual or guidance information. When the prompt upgrades to Level 1, which distinguishes between single-turn and multiple-turn expression types, the BLEU score improves slightly (0.51). However, a more significant increase occurred at Level 2, when the translation style of contextualisation was added to the prompt, which increased the BLEU score to 40.25. This suggests that the inclusion of the translation style has a significant positive impact on the quality of the translation compared to the base translation requirement. At Level 3, the integration of lexical information into the translation prompt further increased the BLEU score to 41.2571. This result suggests that the introduction of lexical information can provide the model with more precise linguistic information, which helps to generate more accurate translations. Finally, the Level 4 prompt achieved the highest BLEU score of 42.88 by including clear instructions, explicit information about the translation context, few-shot examples, and asking the model to check and revise the translation results. Compared to Level 3, the increase was 1.63 points. This largest increase fully demonstrates the remarkable effectiveness of the T3S taxonomy in guiding high-quality translation. Importantly, the Level 4 translation quality is higher than the zero-shot translation quality (BLEU score of 42.50) of GPT-4 under the same dataset and evaluation metrics \cite{jiao2023chatgpt}. Moreover, the results for the ChrF, ROUGE series (F1 average), and TER mainly support the above findings.

Overall, the experimental results consistently show that the translation performance of ChatGPT improves significantly as the level of the prompts increases. This suggests that the T3S taxonomy is not only reasonable but also effective in providing targeted guidance for ChatGPT's translation tasks. Future work can explore the application of the T3S taxonomy to more language pairs, as well as the applicability and optimisation space in different domains and scenarios.

\section{Case Analysis}

This section presents how ChatGPT's performance under the same translation task can be assessed by our T3S taxonomy through a specific translation case. We demonstrate the translation example with different levels of prompts below.

\textbf{Example Text:}
MacBook Air is all you — pick your size, pick your color, then go. Whichever model you choose, it’s built with the planet in mind, with a durable 100 percent recycled aluminum enclosure. And a fanless design means it stays silent even under intense workloads.

\textbf{Expert Translation:} 
\begin{CJK*}{UTF8}{gbsn}
MacBook Air尽显你的风格，挑个合你的款，选个配你的色，直接搞定。两款机型都采用100\%再生铝金属外壳，不管选哪款，都能体现对地球的心意。而无风扇设计意味着即使工作强度高，也能保持静谧无声。\\
\end{CJK*}

\textbf{Level 0 Prompt:}
\textit{Please translate the following text into Chinese:} MacBook Air is all you — pick your size, pick your color, then go. Whichever model you choose, it’s built with the planet in mind, with a durable 100 percent recycled aluminum enclosure. And a fanless design means it stays silent even under intense workloads.

\textbf{Level 0 Translation:}
\begin{CJK*}{UTF8}{gbsn}
MacBook Air 是完全为您打造的 — 选择您的尺寸，选择您的颜色，然后开始使用。无论您选择哪个型号，它都是考虑地球环保的，采用了耐用的100\%回收铝外壳。而且无风扇设计意味着即使在高强度工作下，它也能保持静音。\\
\end{CJK*}
\textbf{Level 1 Prompt (Multiple-Turn form):}
\textit{First turn: Please translate the following text into Chinese:} MacBook Air is all you — pick your size, pick your color, then go. Whichever model you choose, it’s built with the planet in mind, with a durable 100 percent recycled aluminum enclosure. And a fanless design means it stays silent even under intense workloads.
\textit{Second turn: Please check and revise the translation results.}

\textbf{Level 1 Translation:} 
\begin{CJK*}{UTF8}{gbsn}
MacBook Air就是为你而设计的 — 选择你的尺寸，选择你的颜色，然后开始使用。无论你选择哪个型号，它都是为了地球着想，采用了耐用的100\%回收铝外壳。而且，它的无风扇设计意味着即使在高负荷工作下也能保持宁静。\\
\end{CJK*}
\textbf{Level 2 Prompt (Advertising style):}
\textit{Please translate the following text into Chinese in a concise, impressive and advertising style:} MacBook Air is all you — pick your size, pick your color, then go. Whichever model you choose, it’s built with the planet in mind, with a durable 100 percent recycled aluminum enclosure. And a fanless design means it stays silent even under intense workloads.

\textbf{Level 2 Translation:} 
\begin{CJK*}{UTF8}{gbsn}
MacBook Air全由您决定 — 选择您的尺寸、挑选您的颜色，即刻出发。无论您选择哪个型号，它都以地球为念，采用坚固的100\%再生铝外壳打造。无风扇设计确保在高强度工作中也能保持绝对宁静。\\
\end{CJK*}
\textbf{Level 3 Prompt (with POS information):} 
\textit{Given the context of an advertisement for an electronic product and the POS tags, please translate this specific sentence into Chinese in a concise, impressive and advertising style:} MacBook Air (Noun) is (Verb) all you (Pronoun) — pick (Verb) your (Pronoun) size (Noun), pick (Verb) your (Pronoun) color (Noun), then (Adverb) go (Verb). Whichever (Determiner) model (Noun) you (Pronoun) choose (Verb), it’s (Verb) built (Verb) with (Preposition) the (Determiner) planet (Noun) in (Preposition) mind (Noun), with (Preposition) a (Determiner) durable (Adjective) 100 percent (Percentage) recycled (Verb) aluminum (Noun) enclosure (Noun). And (Conjunction) a (Determiner) fanless (Adjective) design (Noun) means (Verb) it (Pronoun) stays (Verb) silent (Adjective) even (Adverb) under (Preposition) intense (Adjective) workloads (Noun).

\textbf{Level 3 Translation:} 
\begin{CJK*}{UTF8}{gbsn}
MacBook Air就是您的综合选择 — 选择您的理想尺寸，选取心仪的颜色，然后出发。无论您选择哪个型号，它都以地球为念，采用坚固的百分之百再生铝质外壳。而无风扇设计确保即使在高强度工作负荷下，也能保持绝对宁静。\\
\end{CJK*}
\textbf{Level 4 Prompt (with Few-shot examples):} 
\textit{First turn: Context Information: It is extracted from an advertisement for an electronic product.
Few-shot Examples:
1.	Translate “Two perfect sizes. Whether you pick the 13- or 15‑inch laptop, both models are superlight and measure just under half an inch thin, so you can take yours anywhere.” into
\begin{CJK*}{UTF8}{gbsn}
“两个尺寸两相宜。13英寸和15英寸两款笔电都十分轻巧纤薄，厚度仅有1厘米多，选哪款都称心称手，去哪里都自由自在。”
\end{CJK*}
2.	Translate “Four stellar colors. Your options are out of this world — and each one comes with a matching MagSafe charging cable.” into 
\begin{CJK*}{UTF8}{gbsn}
“四款配色都惹眼。每种选择都靓出天际，还配有同色系的MagSafe磁吸充电线。”
\end{CJK*}
Considering the context information, few-shot examples and POS tags, please translate the following text into Chinese in a concise, impressive and advertising style:} MacBook Air (Noun) is (Verb) all you (Pronoun) — pick (Verb) your (Pronoun) size (Noun), pick (Verb) your (Pronoun) color (Noun), then (Adverb) go (Verb). Whichever (Determiner) model (Noun) you (Pronoun) choose (Verb), it’s (Verb) built (Verb) with (Preposition) the (Determiner) planet (Noun) in (Preposition) mind (Noun), with (Preposition) a (Determiner) durable (Adjective) 100 percent (Percentage) recycled (Verb) aluminum (Noun) enclosure (Noun). And (Conjunction) a (Determiner) fanless (Adjective) design (Noun) means (Verb) it (Pronoun) stays (Verb) silent (Adjective) even (Adverb) under (Preposition) intense (Adjective) workloads (Noun).
\textit{Second turn: Please check and proofread the translation to ensure that no errors have been made.}

\textbf{Level 4 Translation:}
\begin{CJK*}{UTF8}{gbsn}
MacBook Air 轻薄本，是你的不二之选——选尺寸、选颜色，然后出发吧！无论您选择哪种型号，它都专为地球环保而设计，采用100\%可回收的铝合金外壳，坚固耐用。而且，无风扇设计确保即使它在高强度工作下也能保持宁静。\\
\end{CJK*}
As evident from the foregoing, Level 4 provides the closest results to the expert translation. To further validate the eﬀicacy of this taxonomy, we set up both human and machine evaluation methods.

1)	We invite three proficient translators to evaluate these five levels of translations according to the quantitative assessment criteria for translation quality proposed by \namecite{lixinhua1991}, employing the expert translation as the reference standard. This assessment method was set up with four scoring metrics, namely accuracy, fluency, style and textual coherence, with the following weights: $W_A$=3.5,$W_F$=2.5,$W_S$=2,$W_C$=2. The source text is an advertisement text, which aims at conveying specific messages, emotions and appeals to attract the attention and resonance of the target audience, thus making accuracy and  fluency the primary principles of advertisement translation \cite{xiao2010}. Advertising's main objective is to promote a product or service, rather than a literature. While style and textual coherence can enhance the appeal and taste of an advert, it should not normally come at the expense of accuracy and fluency. Hence, for this case, the weighting cited as such can provide reasonable constraints on the role played by the scoring indicators in terms of importance and priority.

More specifically, accuracy involves the key purpose of translation, which is to ensure that the translation accurately conveys the message and meaning expressed in the original text. Fluency, on the other hand, emphasises the quality of the written expression of the translation, including regularity, clarity and linguistic fluency. Style stresses the importance of appropriately conveying the stylistic, social and local characteristics of the original text, as well as the extent of the use of rhetorical devices, while maintaining the accuracy of the actual meaning. At last, textual coherence considers whether the arrangement of utterances in the translation adequately takes into account the primary and secondary relationships of the information in the original text, whether contextual co-ordination is achieved, and whether coherence of tone is maintained \cite{lixinhua1991}.

Each criterion had a maximum attainable score of ten. The ensuing scores represent the respective assessments of the aforementioned trio of professional translators.  Moreover, with a view to ensuring the reliability of the results and presenting the evaluation results more explicitly, we calculated the final weighted scores using the average scores of the three translators for all levels of translations under different factors.

As shown in Equation \ref{gs1} and Table \ref{table2}, we show the calculation method for the final scores, as well as the different scores and final scores for each level of translation in terms of accuracy, fluency, style, and textual coherence. In the Equation \ref{gs1}, $n$ stands for the number of professional translators, $n$=3; $W_A$, $W_F$, $W_S$, $W_C$ represent the different weights of the four indicators; $A_i$ stands for the specific translator's score for each level of translation in terms of accuracy, $F_i$ stands for the specific translator's score for each level of translation in terms of fluency, $S_i$ refers to the specific translator's score for each level of translation in terms of Style, and Ci denotes the specific translator's score for each level of translation in terms of textual Coherence, $i$=1, 2, 3.

% \usepackage{multirow}
% \usepackage{graphicx}
\begin{table*}[]
\begin{center}
\renewcommand{\arraystretch}{1.5}
\begin{tabular}{|c|ccc|ccc|ccc|ccc|c|}
\hline
\multirow{2}{*}{} & \multicolumn{3}{c|}{Accuracy}                           & \multicolumn{3}{c|}{Fluency}                             & \multicolumn{3}{c|}{Style}                             & \multicolumn{3}{c|}{Coherence}                         & \multirow{2}{*}{Final Score} \\ \cline{2-13}
                  & \multicolumn{1}{c|}{T1} & \multicolumn{1}{c|}{T2} & T3 & \multicolumn{1}{c|}{T1} & \multicolumn{1}{c|}{T2} & T3 & \multicolumn{1}{c|}{T1} & \multicolumn{1}{c|}{T2} & T3 & \multicolumn{1}{c|}{T1} & \multicolumn{1}{c|}{T2} & T3 &                              \\ \hline
Level 0           & \multicolumn{1}{c|}{7}  & \multicolumn{1}{c|}{6}  & 7   & \multicolumn{1}{c|}{7}   & \multicolumn{1}{c|}{7}   & 7  & \multicolumn{1}{c|}{6}  & \multicolumn{1}{c|}{7}  & 6  & \multicolumn{1}{c|}{7}  & \multicolumn{1}{c|}{8}  & 7  & 6.8                          \\ \hline
Level 1           & \multicolumn{1}{c|}{7}  & \multicolumn{1}{c|}{7}  & 7   & \multicolumn{1}{c|}{8}   & \multicolumn{1}{c|}{8}   & 7  & \multicolumn{1}{c|}{7}  & \multicolumn{1}{c|}{7}  & 7  & \multicolumn{1}{c|}{8}  & \multicolumn{1}{c|}{8}  & 7  & 7.3                          \\ \hline
Level 2           & \multicolumn{1}{c|}{9}  & \multicolumn{1}{c|}{8}  & 8   & \multicolumn{1}{c|}{8}   & \multicolumn{1}{c|}{9}   & 8  & \multicolumn{1}{c|}{8}  & \multicolumn{1}{c|}{8}  & 8  & \multicolumn{1}{c|}{8}  & \multicolumn{1}{c|}{9}  & 9  & 8.0                          \\ \hline
Level 3           & \multicolumn{1}{c|}{8}  & \multicolumn{1}{c|}{9}  & 8   & \multicolumn{1}{c|}{9}   & \multicolumn{1}{c|}{7}   & 9  & \multicolumn{1}{c|}{8}  & \multicolumn{1}{c|}{7}  & 7  & \multicolumn{1}{c|}{9}  & \multicolumn{1}{c|}{9}  & 9  & 8.2                          \\ \hline
Level 4           & \multicolumn{1}{c|}{9}  & \multicolumn{1}{c|}{9}  & 8   & \multicolumn{1}{c|}{9}   & \multicolumn{1}{c|}{8}   & 9  & \multicolumn{1}{c|}{9}  & \multicolumn{1}{c|}{8}  & 9  & \multicolumn{1}{c|}{9}  & \multicolumn{1}{c|}{9}  & 9  & 8.8                          \\ \hline
\end{tabular}%

\caption{Assessment of Accuracy, Fluency, Style and Coherence of Translations at All Levels. \protect\\}
\label{table2}
\end{center}

\end{table*}

\begin{figure*}[!htpb]

\begin{equation}
FinalScore=\frac{\sum_{i=1}^n\left(W_A * A_i+W_F * F_i+W_S * S_i+W_C * C_i\right)}{n}
\end{equation}

\end{figure*}
% \begin{equation}
% \begin{split}
% F = \{F_{x} \in  F_{c} &: (|S| > |C|) \\
% &\quad \cap (\text{minPixels} < |S| < \text{maxPixels}) \\
% &\quad \cap (|S_{\text{conected}}| > |S| - \epsilon) \}
% \end{split}
% \end{equation}

Based on the score data presented in Table \ref{table2}, it can be clearly observed that the progression from Level 0 to Level 4 is marked by a discernible trend toward translations that exhibit a greater proximity to expert translation. Specifically, the lowest rated Level 0 (6.8) and Level 1 (7.3) translations score approximately the same in terms of accuracy, and differs significantly in terms of fluency, style and textual coherence. This is due to the fact that their prompts are only different in expression types. Moreover, the improvement (1.0; 0.7; 0.4) in fluency, style and textual coherence from Level 0 to Level 1 proves, to some extent, the effectiveness of multiple-turn prompts in improving the quality of ChatGPT translation results. Compared to Level 2, Level 3 has only a slight improvement (0.2), which is due to the fact that the lexical nature of the source text does not confuse ChatGPT. However, specific texts like legal documents, medical literature, and technical documents often contain specialized terminology, intricate grammatical structures, and polysemous words, and demand a high level of precision and professionalism. In such cases, the inclusion of POS tags becomes crucial as they furnish essential grammatical and semantic information necessary for effectively processing these texts. It should be noted that the addition of POS tags may also increase the preprocessing workload, thus requiring a comprehensive consideration of task requirements and efficiency. Finally, the highest-scoring Level 4 (8.8) translation results exemplify the importance of few-shot examples, which enable ChatGPT to understand the task requirements as well as possible and generate the most brand-specific translations.

2)	LLMs not only show excellent capabilities in several NLP tasks such as machine translation, text summarisation, etc., but they are also state-of-the-art translation quality evaluators \cite{kocmi-federmann-2023-large}. \namecite{kocmi-federmann-2023-large} proposed a GPT-based metric for translation quality assessment, namely GPT Estimation Metric Based Assessment (GEMBA). By conducting experiments on nine versions of GPT models, including ChatGPT and GPT-4, they demonstrated the usefulness and accuracy of pre-trained generative LLMs for translation quality assessment at the system level by using a zero-shot standard prompt. However, this prompt performed poorly at the segment level. Subsequently, \namecite{lu2023error} further validated the capability of LLMs in assessing machine-translated translations. They combined the Chain-of-Thought (CoT) prompting strategy \cite{wei2022chain} and the Error Analysis (EA) paradigm \cite{lu-etal-2022-toward} to propose a novel prompting strategy, Error Analysis Prompting (EAPrompt). EAPrompt divides the scoring process into two stages: first, the LLM is prompted to identify the major and minor errors in the translation. The LLM is then asked to count the number of errors in both categories and calculate the final score. Unlike standard prompts, EAPrompt can produce human-like evaluations of machine translations at both the system and segment levels.

Based on such a pioneering discovery, we perform evaluations of all levels of translations in this case with EAPrompt. Below we show the specific prompt template and the scores for each level of translation. Due to space constraints, the Q\&A sessions for each level of translation are not presented.

\textbf{First-turn: }\\
\textit{(Source Text)\\
  (Reference)\\
  (Translation)\\
  Based on the given source and reference, identify the major and minor errors in this translation. Note that Major errors refer to actual translation or grammatical errors, and Minor errors refer to smaller imperfections, and purely subjective opinions about the translation.}

\textbf{Second-turn: }\\
\textit{Count the number of major and minor errors identified in your last response and compute the final score for this translation. Deduct 5 points for each major error. Deduct 1 point for each minor error. If the translation has no errors, its score will be 0.}

\begin{table}[H]
\renewcommand{\arraystretch}{1.5} % 调整行高
\resizebox{\columnwidth}{!}{%
\begin{tabular}{|l|l|l|l|l|l|}
\hline
        & Level 0 & Level 1 & Level 2 & Level 3 & Level 4 \\ 
\hline
Results & \multicolumn{1}{c|}{-27} & \multicolumn{1}{c|}{-23} & \multicolumn{1}{c|}{-22} & \multicolumn{1}{c|}{-18} & \multicolumn{1}{c|}{-12} \\ 
\hline
\end{tabular}%
}
\caption{Results of ChatGPT's Quality Assessment of Five Levels of Translations under EAPrompt.}
\label{Table3}
\end{table}

Based on Table \ref{Table3}, it can be concluded that the translation quality exhibits an upward trend as the prompt level increases, which is roughly in line with the results of the human-based translation quality assessment. This result further confirms the effectiveness of T3S Taxonomy and the potential of LLMs in translation quality assessment. However, it is also noted that even higher quality translations (e.g. Level 4 Translation) still received negative scores. This suggests that ChatGPT using EAPrompt may have some rigour in the assessment process or be highly sensitive to subtle differences in translations. This could be due to the fact that the LLMs can capture subtle semantic differences and expressive inconsistencies that may seem acceptable to a human evaluator.

\section{Conclusions and Future Directions}

This paper highlighted the significance of a taxonomy of prompts for translation tasks, identifying critical design elements such as expression type, style, POS tagging, and few-shot examples. Furthermore, we explored in detail the key roles of gradable translation prompting taxonomy with explicit descriptions and contextual information to enhance the prompts' quality. The synergistic effect of these factors helps to improve translation quality and avoid misunderstanding and ambiguity, which in turn provides more precise guidance for the ChatGPT translation task. Based on the above, we conducted the T3S taxonomy of prompts for ChatGPT translation tasks.

In our study, we evaluated the effectiveness of our translation taxonomy by conducting an experiment using open-source datasets and standard evaluation metrics to rate translation quality across five levels. We also showcased the taxonomy's usefulness through a case study with ChatGPT, highlighting how prompt design impacts translation performance. Our findings offer valuable insights for enhancing ChatGPT's translation applications, prompt optimization, and overall translation quality and efficiency. Building on the insights gathered from our current investigation, future research could focus on comparing ChatGPT, prompted with our refined taxonomy, with dedicated translation services like Google Translate. This would assess whether our taxonomy-based approach can improve the translation capabilities of LLMs to outperform established translation services.

\section{Acknowledgements}
This work was supported by Gongdong Provincial Key Laboratory of Novel Security Intelligence Technologies (Grant No.2022B1212010005) and XJTLU Research Development Funding (Grant No.RDF-22-01-053).

\bibliographystyle{fullname}
\bibliography{EjemploARTsepln}

\end{document}